\documentclass[11pt,a4paper]{article}
\usepackage[hyperref]{emnlp2018}
\usepackage{times}
\usepackage{latexsym}

\usepackage{url}

\usepackage[utf8]{inputenc} 
\usepackage[T1]{fontenc}    
\usepackage{hyperref}       
\usepackage{url}            
\usepackage{booktabs}       
\usepackage{amsfonts}       
\usepackage{nicefrac}       
\usepackage{microtype}      
\usepackage{booktabs}
\usepackage{tablefootnote}
\usepackage{graphicx}
\usepackage{multirow}
\usepackage{color}
\usepackage{subcaption}
\usepackage{amsmath}
\usepackage{appendix}

\newcommand\norm[1]{\left\lVert#1\right\rVert}

\aclfinalcopy 

\title{Coherence-Aware Neural Topic Modeling}

\author{Ran Ding, Ramesh Nallapati, Bing Xiang\\
  Amazon Web Services \\
  {\tt \{rding, rnallapa, bxiang\}@amazon.com}}

\date{}

\begin{document}

\maketitle
\begin{abstract}
Topic models are evaluated based on their ability to describe documents well (i.e. low perplexity) and to produce topics that carry coherent semantic meaning. In topic modeling so far, perplexity is a direct optimization target. However, topic coherence, owing to its challenging computation, is not optimized for and is only evaluated after training. In this work, under a neural variational inference framework, we propose methods to incorporate a topic coherence objective into the training process. We demonstrate that such a coherence-aware topic model exhibits a similar level of perplexity as baseline models but achieves substantially higher topic coherence.
\end{abstract}

\section{Introduction}

In the setting of a topic model \cite{blei2012probabilistic}, perplexity measures the model's capability to describe documents according to a generative process based on the learned set of topics. In addition to describing documents well (i.e. achieving low perplexity), it is desirable to have topics (represented by top-$N$ most probable words) that are human-interpretable. Topic interpretability or coherence can be measured by \emph{normalized point-wise mutual information} (NPMI) \cite{aletras2013evaluating, lau2014machine}. The calculation of NPMI however is based on look-up operations in a large reference corpus and therefore is non-differentiable and computationally intensive. Likely due to these reasons, topic models so far have been solely optimizing for perplexity, and topic coherence is only evaluated after training. As has been noted in several publications \cite{chang2009reading}, optimization for perplexity alone tends to negatively impact topic coherence. Thus, without introducing topic coherence as a training objective, topic modeling likely produces sub-optimal results. 

Compared to classical methods, such as mean-field approximation \cite{hoffman2010online} and collapsed Gibbs sampling \cite{griffiths2004finding} for the latent Dirichlet allocation (LDA) \cite{blei2003latent} model, neural variational inference \cite{kingma2013auto, rezende2014stochastic} offers a flexible framework to accommodate more expressive topic models. We build upon the line of work on topic modeling using neural variational inference \cite{miao2016neural, miao2017discovering, srivastava2017autoencoding} and incorporate topic coherence awareness into topic modeling. 

Our approaches of constructing topic coherence training objective leverage pre-trained word embeddings \cite{mikolov2013efficient, pennington2014glove, salle2016matrix, joulin2016bag}. The main motivation is that word embeddings carry contextual similarity information that is highly related to the mutual information terms involved in the calculation of NPMI. In this paper, we explore two methods: (1) we explicitly construct a differentiable surrogate topic coherence regularization term; (2) we use word embedding matrix as a factorization constraint on the topical word distribution matrix that implicitly encourages topic coherence.

\section{Models}

\subsection{Baseline: Neural Topic Model (NTM)}

\begin{figure}
  \centering
  \includegraphics[width=0.45\textwidth]{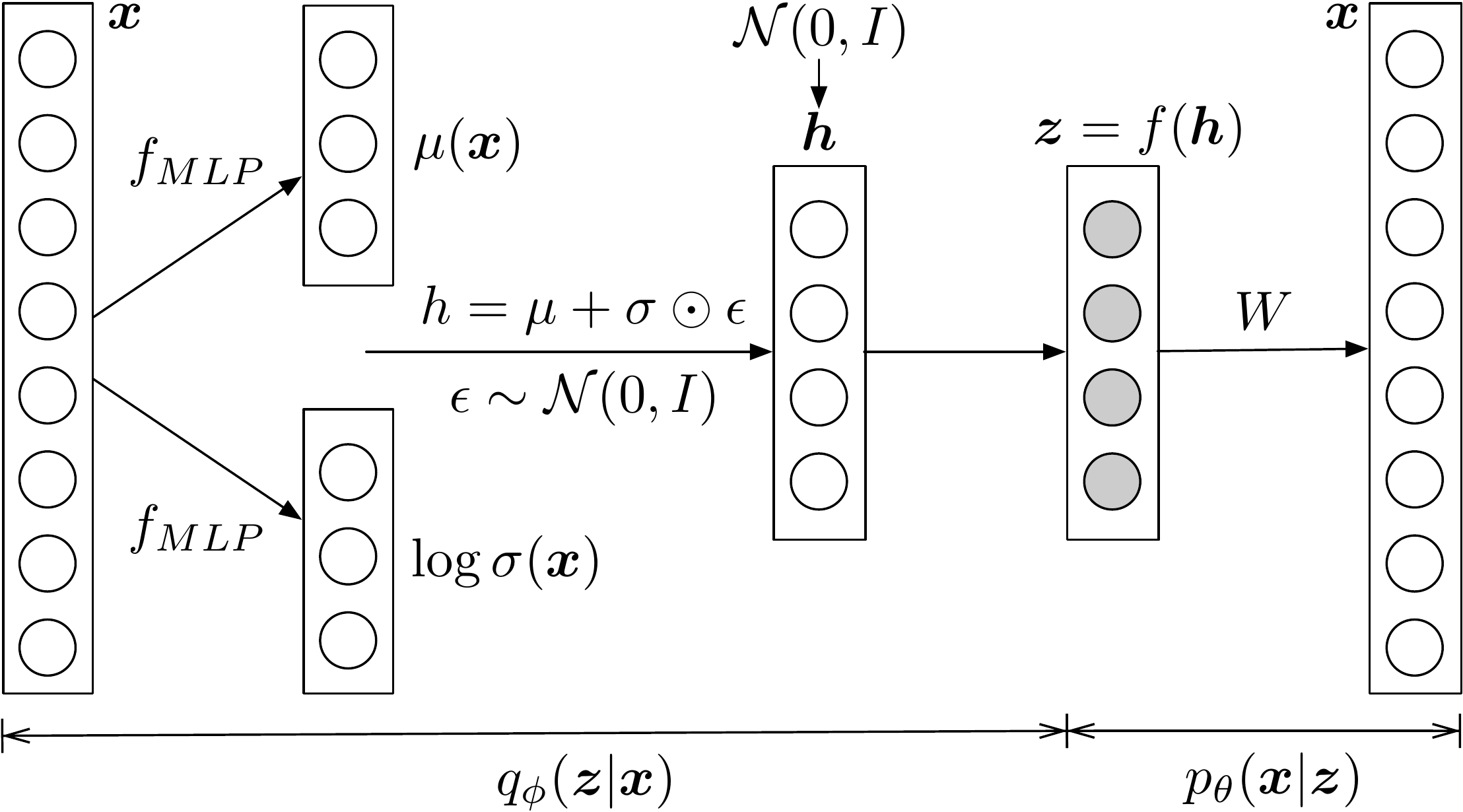}
  \caption{Model architecture} 
  \label{fig:ntm}
\end{figure}

The model architecture shown in Figure \ref{fig:ntm} is a variant of the Neural Variational Document Model (NVDM) \cite{miao2016neural}. Let $x \in \mathbb{R}^{|V| \times 1}$ be the bag-of-words (BOW) representation of a document, where $|V|$ is the size of the vocabulary and let $z \in \mathbb{R}^{K \times 1}$ be the latent topic variable, where $K$ is the number of topics. In the encoder $q_\phi(z|x)$, we have $\pi = f_{MLP}(x)$, $\mu(x) = l_1(\pi)$, $\log \sigma(x) = l_2(\pi)$, $h(x, \epsilon) = \mu + \sigma \odot \epsilon$, where $\epsilon \sim \mathcal{N}(0,I)$, and finally $z=f(h)=\mathrm{ReLU}(h)$. The functions $l_1$ and $l_2$ are linear transformations with bias. We choose the multi-layer perceptron (MLP) in the encoder to have two hidden layers with $3 \times K$ and $2 \times K$ hidden units respectively, and we use the sigmoid activation function. The decoder network $p_\theta(x|z)$ first maps $z$ to the predicted probability of each of the word in the vocabulary $y \in \mathbb{R}^{|V| \times 1}$ through $y=\text{softmax}(Wz+b)$, where $W \in \mathbb{R}^{|V| \times K}$. The log-likelihood of the document can be written as $\log p_\theta(x|z)=\sum_{i=1}^{|V|} \{ x \odot \log y \}$. We name this model Neural Topic Model (NTM) and use it as our baseline. We use the same encoder MLP configuration for our NVDM implementation and all variants of NTM models used in Section \ref{sec:exp}. In NTM, the objective function to maximize is the usual \emph{evidence lower bound} (ELBO) which can be expressed as 
\begin{equation*}
\begin{split}
& \mathcal{L}_{ELBO}(x^i) \\
& \approx \frac{1}{L}\sum_{l=1}^{L} \log p_\theta(x^i|z^{i,l}) - D_{KL}(q_\phi(h|x)|| p_\theta(h)
\end{split}
\end{equation*}
where $z^{i,l}=\mathrm{ReLU}(h(x^i, \epsilon^{l}))$, $\epsilon^l \sim \mathcal{N}(0,I)$. We approximate $\mathbb{E}_{z\sim q(z|x)}[\log p_\theta(x|z)]$ with Monte Carlo integration and calculate the Kullback-Liebler (KL) divergence analytically using the fact $D_{KL}(q_\phi(z|x)|| p_\theta(z)) = D_{KL}(q_\phi(h|x)|| p_\theta(h))$ due to the invariance of KL divergence under deterministic mapping between $h$ and $z$.

Compared to NTM, NVDM uses different activation functions and has $z=h$. Miao \shortcite{miao2017discovering} proposed a modification to NVDM called Gaussian Softmax Model (GSM) corresponding to having $z=\text{softmax}(h)$. Srivastava \shortcite{srivastava2017autoencoding} proposed a model called ProdLDA, which uses a Dirichlet prior instead of Gaussian prior for the latent variable $h$. Given a learned $W$, the practice to extract top-$N$ most probable words for each topic is to take the most positive entries in each column of $W$ \cite{miao2016neural, miao2017discovering, srivastava2017autoencoding}. This is an intuitive choice, provided that $z$ is non-negative, which is indeed the case for NTM, GSM and ProdLDA. NVDM, GSM, and ProdLDA are state-of-the-art neural topic models which we will use for comparison in Section \ref{sec:exp}.

\subsection{Topic Coherence Regularization: NTM-R} 

The topic coherence metric NPMI \cite{aletras2013evaluating, lau2014machine} is defined as 
\begin{equation*}
\begin{split}
&\mathrm{NPMI}(\boldsymbol{w}) \\&= \frac{1}{N(N-1)}\sum_{j=2}^{N} \sum_{i=1}^{j-1} \frac{\log\frac{P(w_i, w_j)}{P(w_i)P(w_j)}}{-\log P(w_i, w_j)}
\end{split}
\end{equation*}
where $\boldsymbol{w}$ is the list of top-$N$ words for a topic. $N$ is usually set to 10. For a model generating $K$ topics, the overall NPMI score is an average over all topics. The computational overhead and non-differentiability originate from extracting the co-occurrence frequency from a large corpus\footnote{A typical calculation of NPMI over 50 topics based on the Wikipedia corpus takes $\sim$20 minutes, using code provided by \cite{lau2014machine} at \url{https://github.com/jhlau/topic_interpretability}.}. 

From the NPMI formula, it is clear that word-pairs that co-occur often would score high, unless they are rare word-pairs -- which would be normalized out by the denominator. The NPMI scoring bears remarkable resemblance to the contextual similarity produced by the inner product of word embedding vectors. Along this line of reasoning, we construct a differentiable, computation-efficient word embedding based topic coherence (WETC).

Let $E$ be the row-normalized word embedding matrix for a list of $N$ words, such that $E \in \mathbb{R}^{N \times D}$ and $\norm{E_{i,:}}=1$, where $D$ is the dimension of the embedding space. We can define \emph{pair-wise} word embedding topic coherence in a similar spirit as NPMI: 
\begin{equation*}
\begin{split}
\mathrm{WETC}_{PW}(E) &= \frac{1}{N(N-1)}\sum_{j=2}^{N} \sum_{i=1}^{j-1} \langle E_{i,:}, E_{j,:}\rangle \\
&= \frac{\sum \{E^TE\} - N}{2N(N-1)}
\end{split}
\end{equation*}
where $\langle\cdot,\cdot\rangle$ denotes inner product. Alternatively, we can define \emph{centroid} word embedding topic coherence 

\begin{equation*}
\mathrm{WETC}_{C}(E) =  \frac{1}{N} \sum \{Et^T\}
\end{equation*}
where vector $t \in \mathbb{R}^{1 \times D}$ is the centroid of $E$, normalized to have $\norm{t}=1$. Empirically, we found that the two WETC formulations behave very similarly. In addition, both $\mathrm{WETC}_{PW}$ and $\mathrm{WETC}_{C}$ correlate to human judgement almost equally well as NPMI when using \texttt{GloVe} \cite{pennington2014glove} vectors\footnote{See Appendix A for details on an empirical study of human judgement of topic coherence, NPMI and WETC with various types of word embeddings.}.

With the above observations, we propose the following procedure to construct a WETC-based surrogate topic coherence regularization term: (1) let $E \in \mathbb{R}^{|V| \times D}$ be the pre-trained word embedding matrix for the vocabulary, rows aligned with $W$; (2) form the $W$-weighted centroid (topic) vectors $T \in \mathbb{R}^{D \times K}$ by $T=E^TW$; (3) calculate the cosine similarity matrix $S \in \mathbb{R}^{|V| \times K}$ between word vectors and topic vectors by $S = ET$; (4) calculate the $W$-weighted sum of word-to-topic cosine similarities for each topic $C \in \mathbb{R}^{1 \times K}$ as $C=\sum_i (S\odot W)_{i,:}$. Compared to $\mathrm{WETC}_{C}$, in calculating $C$ we do not perform top-$N$ operation in $W$, but directly use $W$ for weighted sum. Specifically, we use $W$-weighted topic vector construction in Step-2 and $W$-weighted sum of the cosine similarities between word vectors and topic vectors in Step-4. To avoid unbounded optimization, we normalize the rows of $E$ and the columns of $W$ before Step-2, and normalize the columns of $T$ after Step-2. The overall maximization objective function becomes $\mathcal{L}_{R}(x; \theta, \phi) = \mathcal{L}_{ELBO} + \lambda \sum_i{C_i}$, where $\lambda$ is a hyper-parameter with positive values controlling the strength of topic coherence regularization. We name this model NTM-R.

\begin{table}
  \centering
\resizebox{0.45\textwidth}{!}{%

  \begin{tabular}{l|cc|cc}
    \toprule
\multicolumn{1}{r|}{Metric} & \multicolumn{2}{c}{Perplexity} &
\multicolumn{2}{c}{NPMI} \\
    \cmidrule(lr){2-3} 
    \cmidrule(lr){4-5} 
    \multicolumn{1}{r|}{Number of topics} & 50 & 200 & 50 & 200\\
    \midrule
    LDA \\
    \midrule
\quad LDA, mean-field & 1046 & 1195 & 0.11 & 0.06\\ 
\quad LDA, collapsed Gibbs & \textbf{728} &
\textbf{688} & 0.17 & 0.14 \\ 
    \midrule
    Neural Models \\
    \midrule
	\quad NVDM & 750 & 743 & 0.14 & 0.13 \\ 
	\quad GSM & 787 & 829 & 0.22 & 0.19\\ 
	\quad ProdLDA & 1172 & 1168 & 0.28 & 0.24 \\ 
    \quad NTM & 780 & 768 & 0.18 & 0.18 \\ 
	\quad NTM-R & \textcolor{blue}{\underline{\textbf{775}}} & \textcolor{blue}{\underline{\textbf{763}}} & \textcolor{blue}{\underline{\textbf{0.28}}} & \textcolor{blue}{\underline{\textbf{0.23}}} \\ 
    \quad NTM-F & 898 & 1086 & \textbf{0.29} & 0.24 \\ 
    \quad NTM-FR & 924 & 1225 & 0.27 & \textbf{0.26} \\ 
    \bottomrule

 
\end{tabular}
}
\caption{Comparison to LDA and neural variational models on the \emph{20NewsGroup} dataset. Best numbers are bolded. The blue underlined row highlights the best NPMI and perplexity tradeoff as discussed in text.}
\label{table:20ng_table}
\end{table}

\subsection{Word Embedding as a Factorization Constraint: NTM-F and NTM-FR} \label{sec:ntm-f}

Instead of allowing all the elements in $W$ to be freely optimized, we can impose a factorization constraint of $W = E \hat T$, where $E$ is the pre-trained word embedding matrix that is \emph{fixed}, and only $\hat T$ is allowed to be learned through training. Under this configuration, $\hat T$ lives in the embedding space, and each entry in $W$ is the dot product similarity between a topic vector $\hat{T_i}$ and a word vector $E_j$. As one can imagine, similar words would have similar vector representations in $E$ and would have similar weights in each column of $W$. Therefore the factorization constraint encourages words with similar meaning to be selected or de-selected together thus potentially improving  topic coherence.

We name the NTM model with factorization constraint enabled as NTM-F. In addition, we can apply the regularization discussed in the previous section on the resulting matrix $W$ and we name the resulting model NTM-FR.

\section{Experiments and Discussions} \label{sec:exp} 

\begin{figure*}[t]
\centering
\includegraphics[width=0.8\textwidth]{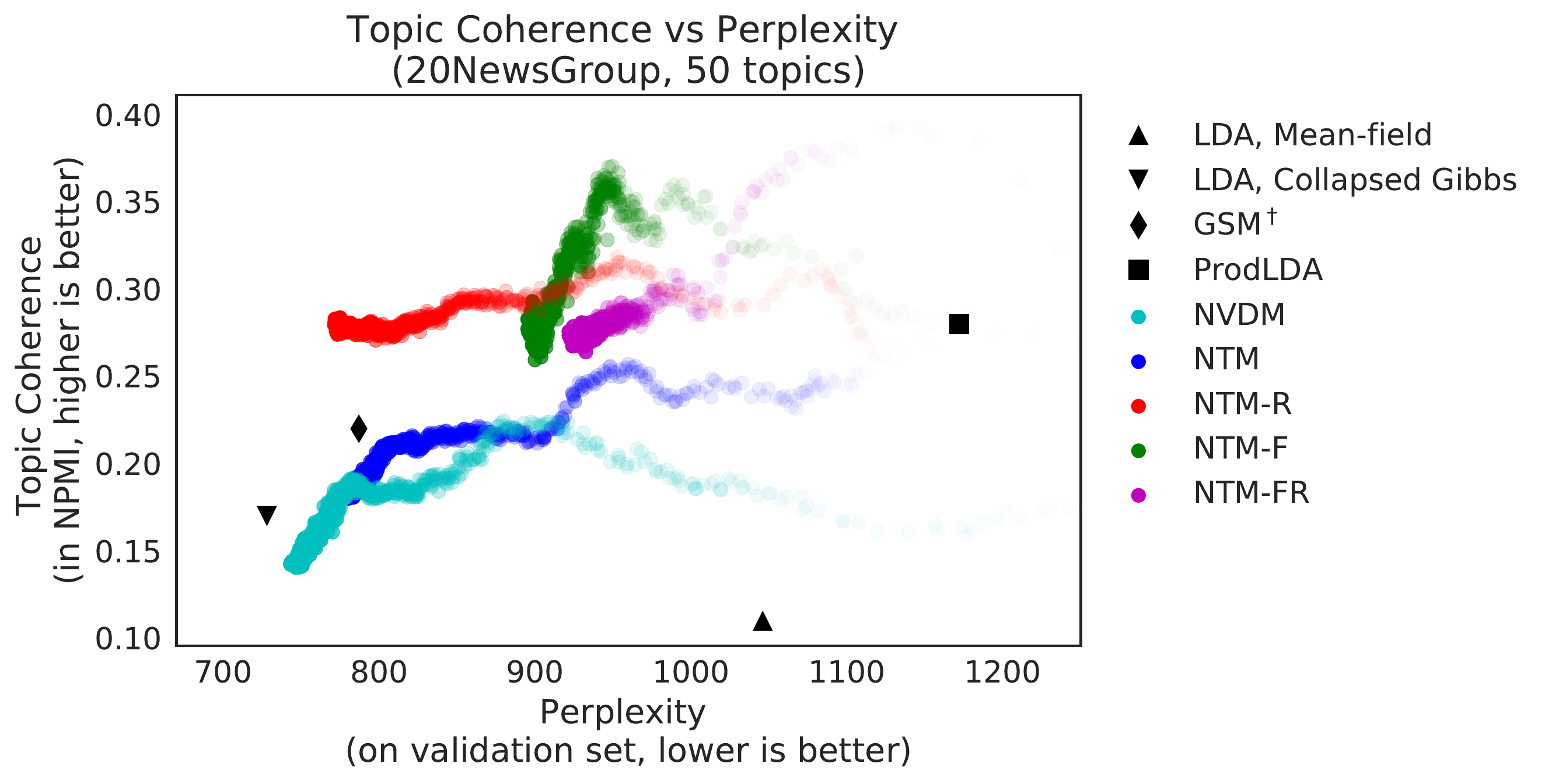}
\caption{NPMI vs. perplexity for various models at 50 topics. For NVDM and NTM models the traces correspond to the evolution over training epochs. High transparency is the beginning of the training.} 
\label{fig:npmi_vs_ppx}
\end{figure*}

\subsection{Results on \emph{20NewsGroup}} 

First, we compare the proposed models to state-of-the-art neural variational inference based topic models in the literature (NVDM, GSM, and ProdLDA) as well as LDA benchmarks, on the \emph{20NewsGroup} dataset\footnote{We use the exact dataset from \cite{srivastava2017autoencoding} to avoid subtle differences in pre-processing}. In training NVDM and all NTM models, we used \texttt{Adadelta} optimizer \cite{zeiler2012adadelta}. We set the learning rate to 0.01 and train with a batch size of 256. For NTM-R, NTM-F and NTM-FR, the word embedding we used is \texttt{GloVe} \cite{pennington2014glove} vectors pre-trained on Wikipedia and Gigaword with 400,000 vocabulary size and 50 embedding dimensions\footnote{Obtained from \url{https://nlp.stanford.edu/projects/glove/}}. The topic coherence regularization coefficient $\lambda$ is set to 50. The results are presented in Table \ref{table:20ng_table}. 

Overall we can see that LDA trained with collapsed Gibbs sampling achieves the best perplexity, while NTM-F and NTM-FR models achieve the best topic coherence (in NPMI). Clearly, there is a trade-off between perplexity and NPMI as identified by other papers. So we constructed Figure \ref{fig:npmi_vs_ppx}, which shows the two metrics from various models. For the models we implemented, we additionally show the full evolution of these two metrics over training epochs.

From Figure \ref{fig:npmi_vs_ppx}, it becomes clear that although ProdLDA exhibits good performance on NPMI, it is achieved at a steep cost of perplexity, while NTM-R achieves similar or better NPMI at much lower perplexity levels. At the other end of the spectrum, if we look for low perplexity, the best numbers among neural variational models are between 750 and 800. In this neighborhood, NTM-R substantially outperforms the GSM, NVDM and NTM baseline models. Therefore, we consider NTM-R the best model overall. Different downstream applications may require different tradeoff points between NPMI and perplexity. However, the proposed NTM-R model does appear to provide tradeoff points on a Pareto front compared to other models across most of the range of perplexity.

\subsection{Comments on NTM-F and NTM-FR}

It is worth noting that although NTM-F and NTM-FR exhibit high NPMI early on, they fail to maintain it during the training process. In addition, both models converged to fairly high perplexity levels. Our hypothesis is that this is caused by NTM-F and NTM-FR's substantially reduced parameter space - from $|V|\times K$ to $D\times K$, where $|V|$ ranges from 1,000 to 150,000 in a typical dataset, while $D$ is on the order of 100. 

Some form of relaxation could alleviate this problem. For example, we can let $W=E\hat T + A$, where $A$ is of size $|V|\times K$ but is heavily regularized, or let $W=EQ\hat T$ where $Q$ is allowed as additional free parameters. We leave fully addressing this to future work.

\subsection{Validation on other Datasets}

\begin{figure*}[t]
    \centering
    \begin{subfigure}[h]{0.3\textwidth}
            \includegraphics[width=\textwidth]{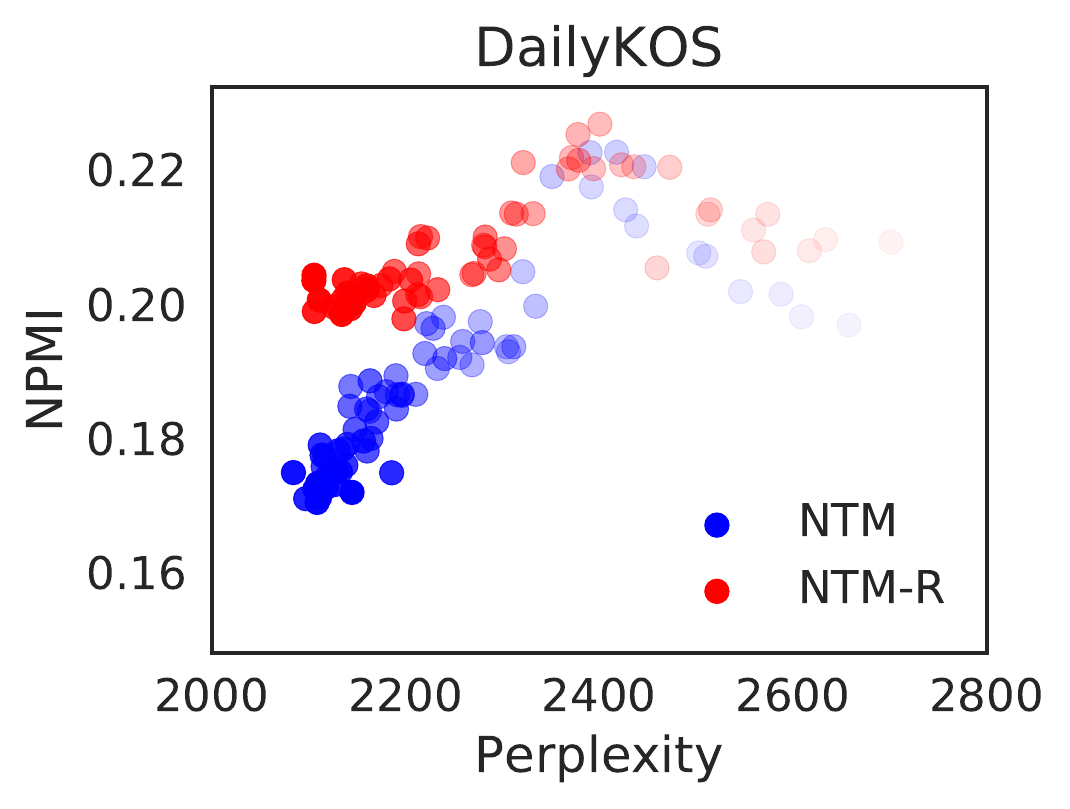}
    \end{subfigure}
    \begin{subfigure}[h]{0.3\textwidth}
            \includegraphics[width=\textwidth]{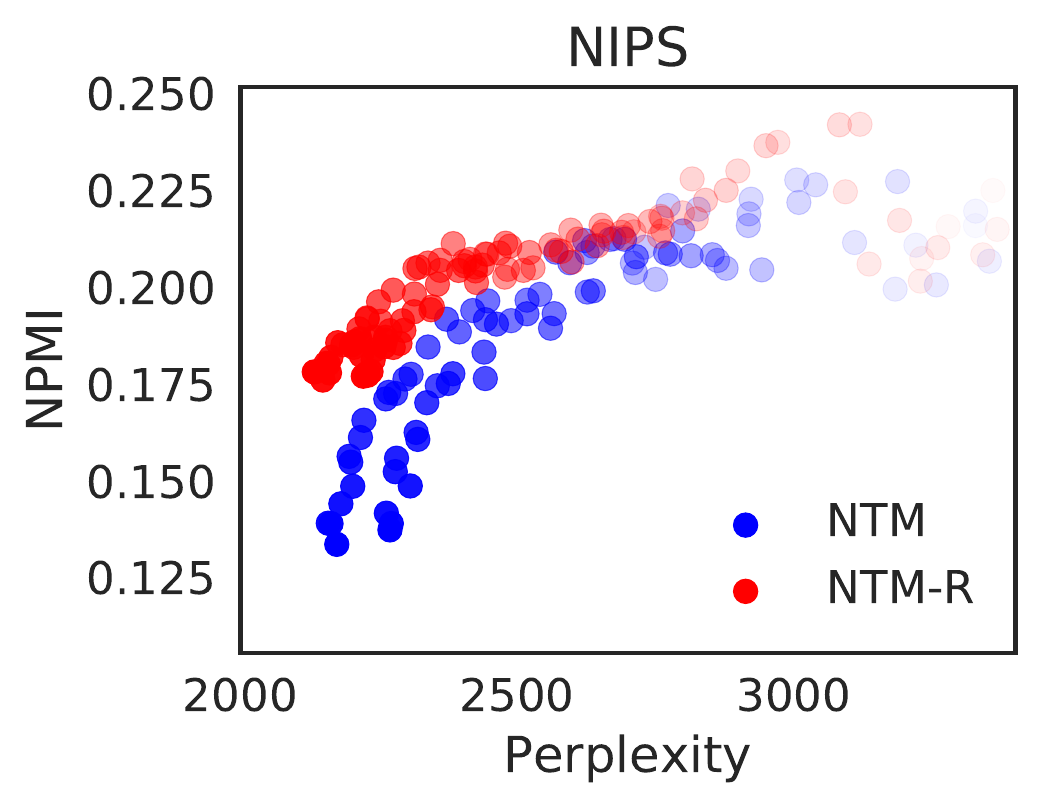}
    \end{subfigure}
    \begin{subfigure}[h]{0.3\textwidth}
            \includegraphics[width=\textwidth]{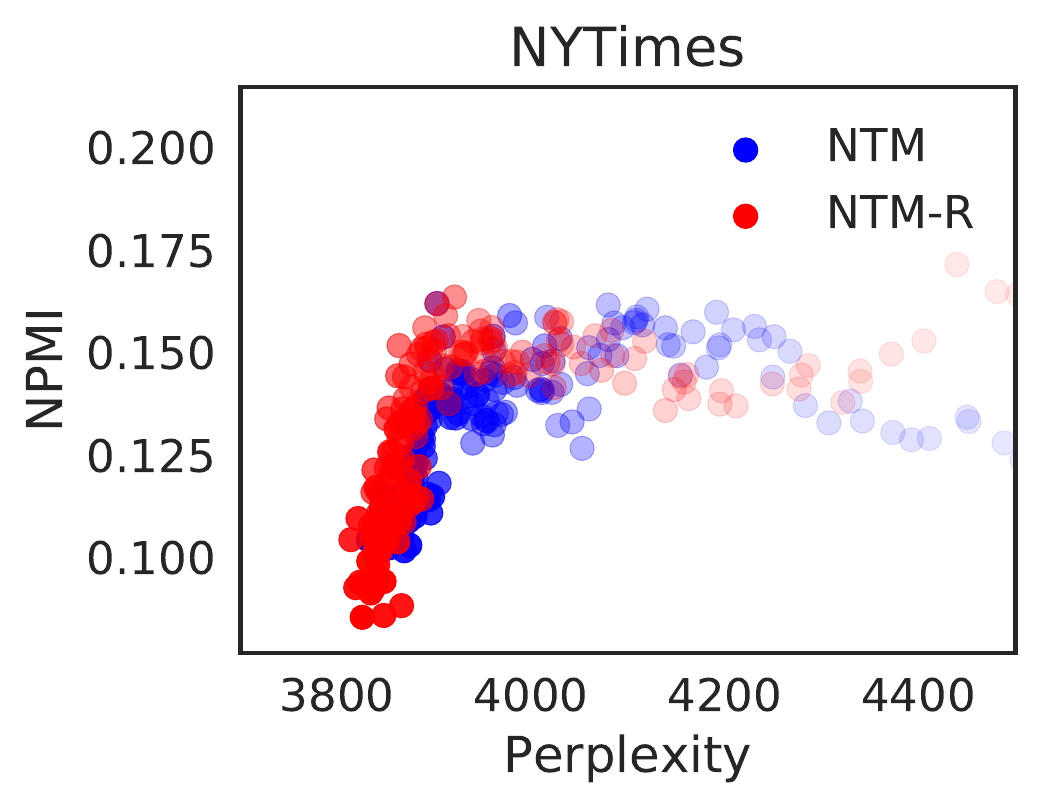}
    \end{subfigure}
    
    \begin{subfigure}[h]{0.3\textwidth}
            \includegraphics[width=\textwidth]{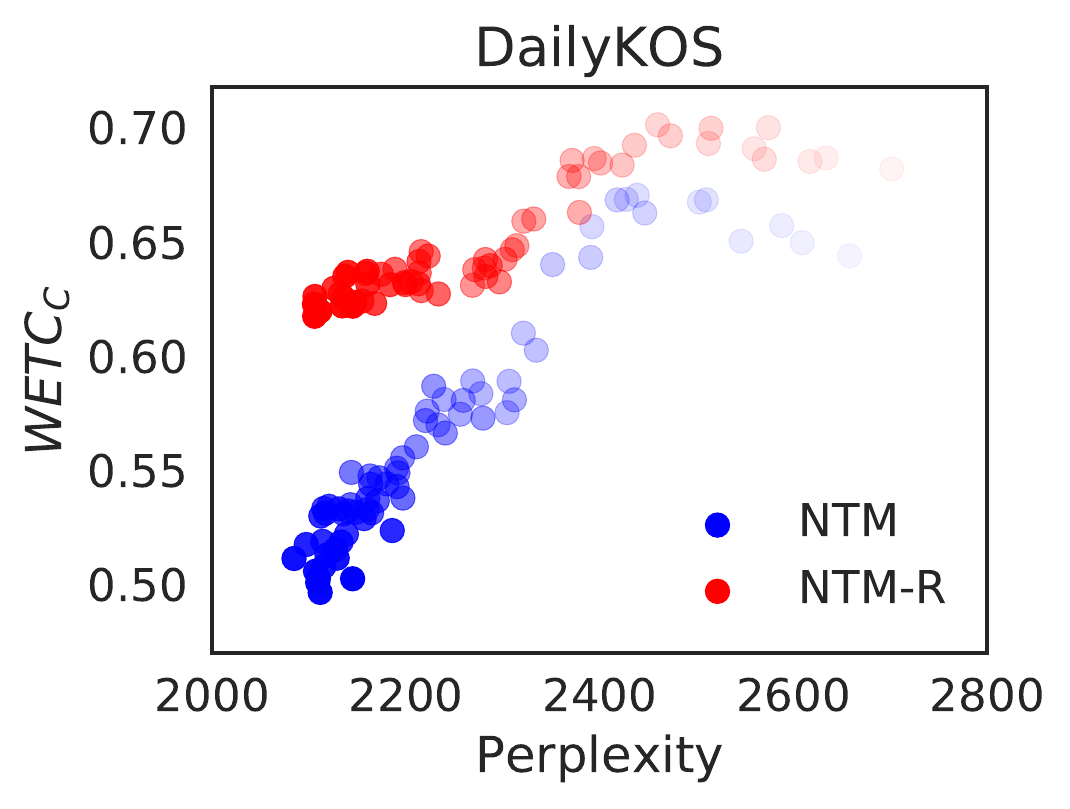}
    \end{subfigure}
    \begin{subfigure}[h]{0.3\textwidth}
            \includegraphics[width=\textwidth]{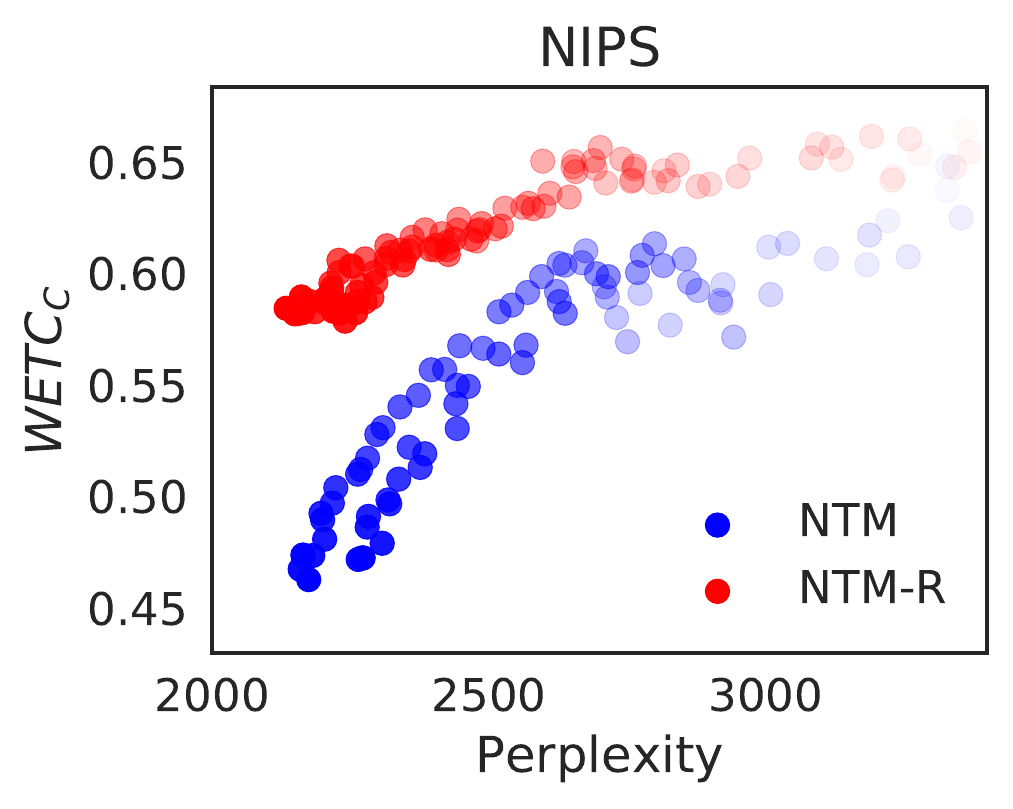}
    \end{subfigure}
    \begin{subfigure}[h]{0.3\textwidth}
            \includegraphics[width=\textwidth]{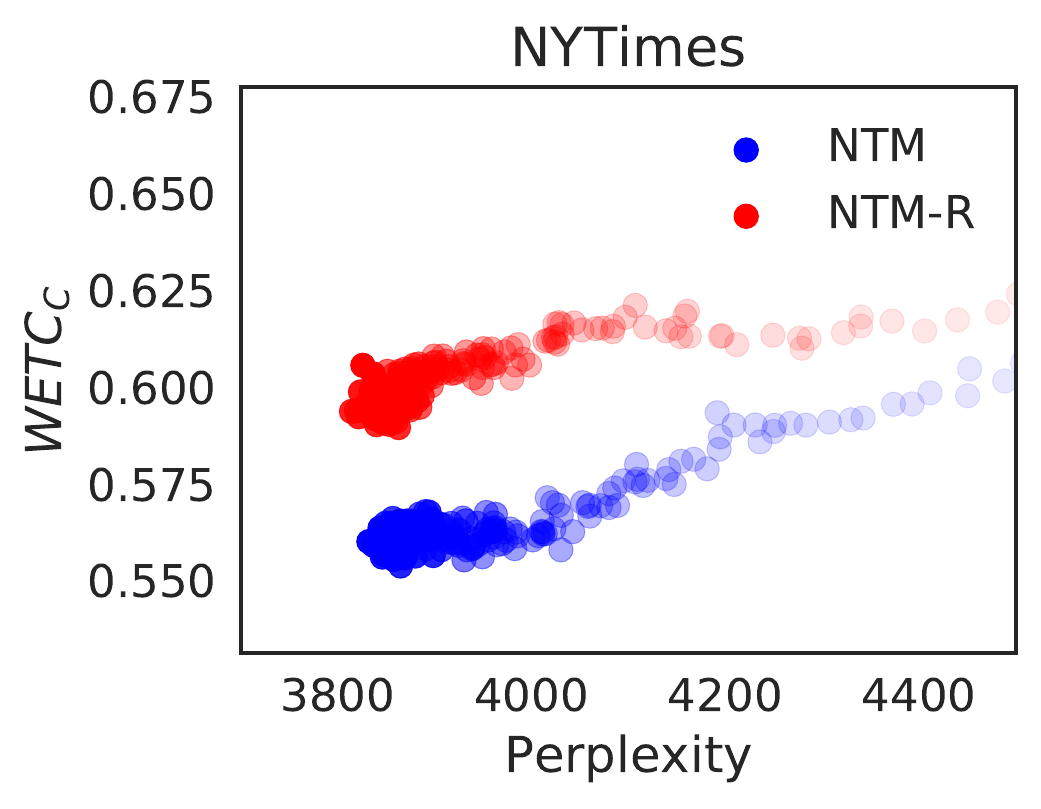}
    \end{subfigure}
    \caption{Performance comparison between NTM-R and NTM on multiple datasets, with 50 topics. Top row is NPMI versus perplexity, bottom row is $\mathrm{WETC}_C$ versus perplexity. From left to right: DailyKOS, NIPS, and NYTimes. See text for details about the datasets.} \label{fig:cross_ds}
\end{figure*}

To further validate the performance improvement from using WETC-based regularization in NTM-R, we compare NTM-R with the NTM baseline model on a few more datasets: DailyKOS, NIPS, and NYTimes\footnote{\url{https://archive.ics.uci.edu/ml/datasets/Bag+of+Words}} \cite{asuncion2007uci}. These datasets offer a wide range of document length (ranging from $\sim$100 to $\sim$1000 words), vocabulary size (ranging from $\sim$7,000 to $\sim$140,000), and type of documents (from news articles to long-form scientific writing). In this set of experiments, we used the same settings and hyperparameter $\lambda$ as before and did not fine-tune for each dataset. The results are presented in Figure \ref{fig:cross_ds}. In a similar style as Figure \ref{fig:npmi_vs_ppx}, we show the evolution of NPMI and WETC versus perplexity over epochs until convergence. 

Among all datasets, we observed improved NPMI at the same perplexity level, validating the effectiveness of the topic coherence regularization. However, on the NYTimes dataset, the improvement is quite marginal even though WETC improvements are very noticeable. One particularity about the NYTimes dataset is that approximately 58,000 words in the 140,000-word vocabulary are named entities. It appears that the large number of named entities resulted in a divergence between NPMI and WETC scoring, which is an issue to address in the future.

%
%

\section{Conclusions}

In this work, we proposed regularization and factorization constraints based approaches to incorporate awareness of topic coherence into the formulation of topic models: NTM-R and NTM-F respectively. We observed that NTM-R substantially improves topic coherence with minimal sacrifice in perplexity. To our best knowledge, NTM-R is the first topic model that is trained with an objective towards topic coherence -- a feature directly contributing to its superior performance. We further showed that the proposed WETC-based regularization method is applicable to a wide range of text datasets.

\bibliography{ref.bib}
\bibliographystyle{acl_natbib_nourl}

\appendix

\section{Word Embedding Topic Coherence} \label{sec:wetc} 

As studied in \cite{aletras2013evaluating} and \cite{lau2014machine}, the NPMI metric for assessing topic coherence over a list of words $\boldsymbol{w}$ is defined in Eq.  \ref{eq:npmi}.
\begin{equation}
\label{eq:npmi}
\begin{split}
&\mathrm{NPMI}(\boldsymbol{w}) \\&= \frac{1}{N(N-1)}\sum_{j=2}^{N} \sum_{i=1}^{j-1} \frac{\log\frac{P(w_i, w_j)}{P(w_i)P(w_j)}}{-\log P(w_i, w_j)}
\end{split}
\end{equation}
where $P(w_i)$ and $P(w_i, w_j)$ are the probability of words and word pairs, calculated based on a reference corpus. $N$ is usually set to 10, by convention, so that NPMI is evaluated over the topic-10 words for each topic. For a model generating $K$ topics, the overall NPMI score is an average over all the topics. The computational overhead comes from extracting the relevant co-occurrence frequency from a large corpus. This problem is exacerbated when the look-up also requires a small sliding window as the authors of \cite{lau2014machine} suggested. A typical calculation of 50 topics based on a few million documents from the Wikipedia corpus takes $\sim$20 minutes\footnote{Using code provided by \cite{lau2014machine} at \url{https://github.com/jhlau/topic_interpretability}. Running parallel processes on 8 Intel Xeon E5-2686 CPUs.}.

For a list of words $\boldsymbol{w}$ of length $N$, we can assemble a corresponding word embedding matrix $E \in \mathbb{R}^{N \times D}$ with each row corresponding to a word in the list. $D$ is the dimension of the embedding space. Averaging across the rows, we can obtain vector $t \in \mathbb{R}^{1 \times D}$ as the centroid of all the word vectors. It may be regarded as a "topic" vector. In addition, we assume that each row of $E$ and $t$ is normalized, i.e. $\norm{t}=1$ and $\norm{E_{i,:}}=1$. With these, we define \emph{pair-wise} and \emph{centroid} word embedding topic coherence $\mathrm{WETC}_{PW}$ and $\mathrm{WETC}_C$ as follows:
\begin{equation}
\label{eq:pw_wetc}
\begin{split}
\mathrm{WETC}_{PW}(E) &= \frac{1}{N(N-1)}\sum_{j=2}^{N} \sum_{i=1}^{j-1} \langle E_{i,:}, E_{j,:}\rangle \\
&= \frac{\sum \{E^TE\} - N}{2N(N-1)}
\end{split}
\end{equation}
\begin{equation}
\label{eq:centroid_wetc}
\mathrm{WETC}_{C}(E) =  \frac{1}{N} \sum \{Et^T\}
\end{equation}
where $\langle\cdot,\cdot\rangle$ denotes inner product. The simplification in Eq. \ref{eq:pw_wetc} is due to the row normalization of $E$. 

In this setting, we have the flexibility to use any pre-trained word embeddings to construct $E$. To experiment, we compared several recently developed variants
\footnote{Details of pre-trained word embeddings used in Table \ref{table:wetc_table}
\begin{itemize}
	\item \texttt{Word2Vec} \cite{mikolov2013efficient}: pre-trained on GoogleNews, with 3 million vocabulary size and 300 embedding dimension. Obtained from \url{https://code.google.com/archive/p/word2vec/}.
	\item \texttt{GloVe} \cite{pennington2014glove}: pre-trained on Wikipedia and Gigaword, with 400,000 vocabulary size and 50 and 300 embedding dimension. Obtained from \url{https://nlp.stanford.edu/projects/glove/}.
	\item \texttt{FastText} \cite{joulin2016bag}: pre-trained on Wikipedia with 2.5 million vocabulary size and 300 embedding dimension. Obtained from \url{https://github.com/facebookresearch/fastText}.
	\item \texttt{LexVec} \cite{salle2016matrix}: pre-trained on Wikipedia with 370,000 vocabulary size and 300 embedding dimension. Obtained from \url{https://github.com/alexandres/lexvec}.
\end{itemize}
}. The dataset from \cite{aletras2013evaluating} provides human ratings for 300 topics coming from 3 corpora: 20NewsGroup (20NG), New York Times (NYT) and genomics scientific articles (Genomics), which we use as the human gold standard. We use Pearson and Spearman correlations to compare NPMI and WETC scores against human ratings. The results are shown in Table \ref{table:wetc_table}.

%
%
%
%
%
%
%
%
%
%
%
%
%
%
%
%
%

\begin{table}[t]
  \centering
\resizebox{0.48\textwidth}{!}{%
  \begin{tabular}{l|cc|cc|cc}
    \toprule
\multicolumn{1}{r|}{Dataset} & \multicolumn{2}{c}{20NG} & \multicolumn{2}{c}{NYT} & \multicolumn{2}{c}{Genomics}\\
    \cmidrule(lr){2-3} 
    \cmidrule(lr){4-5} 
    \cmidrule(lr){6-7} 
\multicolumn{1}{r|}{Correlation} & P & S & P & S & P & S\\
\midrule

\multicolumn{1}{r|}{NPMI} & 0.74 & 0.74 & 0.72 & 0.71 & 0.62 & 0.65 \\

\midrule
GloVe-50d \\
\multicolumn{1}{r|}{$\mathrm{WETC}_{PW}$} & {0.82} & {0.77} & 0.73 & 0.71 & 0.65 & 0.65\\
\multicolumn{1}{r|}{$\mathrm{WETC}_{C}$}  & 0.81 & 0.77 & 0.73 & 0.71 & 0.65 & 0.65\\

\midrule
GloVe-300d \\
\multicolumn{1}{r|}{$\mathrm{WETC}_{PW}$} & 0.77 & 0.75 & 0.78 & 0.76 & {0.68} & {0.70}\\ 
\multicolumn{1}{r|}{$\mathrm{WETC}_{C}$}  & 0.80 & 0.75 & 0.78 & 0.76 & {0.68} & {0.70}\\


\midrule
Word2Vec\\
\multicolumn{1}{r|}{$\mathrm{WETC}_{PW}$} & 0.29 & 0.23 & 0.53 & 0.59 & 0.56 & 0.55\\
\multicolumn{1}{r|}{$\mathrm{WETC}_{C}$}  & 0.31 & 0.23 & 0.55 & 0.59 & 0.56 & 0.55\\

\midrule
FastText\\
\multicolumn{1}{r|}{$\mathrm{WETC}_{PW}$} & 0.40 & 0.61 & 0.63 & 0.67 & 0.62 & 0.62\\
\multicolumn{1}{r|}{$\mathrm{WETC}_{C}$}  & 0.48 & 0.61 & 0.64 & 0.67 & 0.63 & 0.62\\

\midrule
LexVec\\    
\multicolumn{1}{r|}{$\mathrm{WETC}_{PW}$} & 0.37 & 0.57 & 0.79 & 0.80 & 0.65 & 0.64\\
\multicolumn{1}{r|}{$\mathrm{WETC}_{C}$}  & 0.47 & 0.57 & {0.81} & {0.80} & 0.65 & 0.64\\
    
\bottomrule
  \end{tabular}
 }
 \caption{NPMI and WETC correlation with human gold standard (P: Pearson, S: Spearman)}   \label{table:wetc_table}
\end{table}

From Table \ref{table:wetc_table} we observed a minimal difference between pair-wise and centroid based WETC in general. Overall, \texttt{GloVe} appears to perform the best across different types of corpora and its correlation with human ratings is very comparable to NPMI-based scores. Our NPMI calculation is based on the Wikipedia corpus and should serve as a fair comparison. In addition to the good correlation exhibited by WETC, the evaluation of WETC only involves matrix multiplications and summations and thus is fully differentiable and several orders of magnitude faster than NPMI calculations. WETC opens the door of incorporating topic coherence as a training objective, which is the key idea we will investigate in the subsequent sections. It is worth mentioning that, for \texttt{GloVe}, the low dimensional embedding (50d) appears to perform almost equally well as high dimensional embedding (300d). Therefore, we will use Glove-400k-50d in all subsequent experiments.

While the WETC metric on its own might be of interest to the topic modeling research community, we leave the task of formally establishing it as a standard metric in place of NPMI to future work. In this work, we still use the widely accepted NPMI as the objective topic coherence metric for model comparisons. 

\end{document}